%% file: main.tex
\pdfoutput=1

\documentclass[11pt]{article}

\usepackage{EMNLP2023}

\usepackage{times}
\usepackage{latexsym}

\usepackage{booktabs}
\usepackage{graphicx}

\usepackage[T1]{fontenc}

\usepackage[utf8]{inputenc}

\usepackage{microtype}
\usepackage{todonotes}

\usepackage{inconsolata}
\usepackage{listings}
\usepackage{xspace}
\usepackage{lscape,geometry,multicol}
\usepackage{todonotes}
\usepackage{subcaption}
\usepackage{array, multirow}

\newcommand{\STAB}[1]{\begin{tabular}{@{}c@{}}#1\end{tabular}}

\newcommand{\datasetname}{LBPP\xspace} 
\newcommand{\humaneval}{\texttt{HumanEval}\xspace}
\newcommand{\MBPP}{\texttt{MBPP}\xspace}

\title{On Leakage of Code Generation Evaluation Datasets}

\author{{\bf Alexandre Matton}, {\bf Tom Sherborne}, {\bf Dennis Aumiller}, \\
{\bf Elena Tommasone}, {\bf Milad Alizadeh},
{\bf Jingyi He}, \\ {\bf Raymond Ma}, {\bf Maxime Voisin}, {\bf Ellen Gilsenan-McMahon} \\ {\bf Matthias Gall\'e} \\
Cohere
}

\begin{document}
\maketitle
\begin{abstract}
In this paper, we consider contamination by code generation test sets, in particular in their use in modern large language models.
We discuss three possible sources of such contamination and show findings supporting each of them: (i) direct data leakage, (ii) indirect data leakage through the use of synthetic data and (iii) overfitting to evaluation sets during model selection.
To address this, we release \datasetname: an uncontaminated new benchmark of 161 prompts with their associated Python solutions. \datasetname is released at \url{https://huggingface.co/datasets/CohereForAI/lbpp}.
\end{abstract}

\section{Introduction}
\label{sec:intro}
Code generation has emerged as an important skill for large language models to master.
Measuring recent progress in code generation has relied on few, critical benchmarks to judge performance between model families and checkpoints. While many recent sophisticated evaluation datasets have been proposed \citep{jain2024livecodebench,jimenez2024swebench}, the community largely relies on \humaneval~\citep{humaneval} and \MBPP~\citep{mbpp} to judge a new model's code capability. In fact, all major announcements in 2023-2024 claiming advanced code capabilities---from either academic or industry labs---boast one or both of these datasets. Practically, reporting \humaneval and \MBPP is mandatory for a model to report competitive code generation.

However, the importance of these benchmarks has led to a conflict between popularity and utility. 
Obtaining competitive numbers comes with significant scientific and economic reward---made increasingly easy with the proliferation of public replicas and evaluation harnesses featuring these datasets.
However, this prevalence has led to data leakage beyond the original evaluation scope. When this evaluation data \textit{contaminates} model training, the validity of the metrics as a measure of generalization capability becomes unreliable. 
If a model is trained on data used for out-of-distribution generalization (or selected for its performance on that data), we break an implicit tenet of how model capability should be measured. We argue that understanding the effect of contamination is critical to accurately interpreting scores on these benchmarks. 

In this paper, we review the evidence that most contemporary LLMs are \textit{contaminated} with data sourced from these two benchmarks. We define \textit{contamination} here as any procedure leaking datasets \textit{during} model training such that these datasets are no longer unseen at inference. 
The most obvious method of contamination is the presence inside training data. Section \ref{sec:contamination:dataleakage} reviews evidence that these benchmarks are widespread in training corpora in original and paraphrased forms. Unfortunately, it is not feasible to manually remove all the corresponding examples from the training corpora and the most common automatic decontamination methods have low recall. Section \ref{sec:contamination:synth} proposes that contamination also occurs indirectly through the use of synthetic data---a widespread paradigm used to increase coding capabilities by generating additional code training tokens for pre-training or fine-tuning.
Finally, Section \ref{sec:contamination:overfitting} argues that checkpoint selection may be overly influenced by these datasets, overfitting to these benchmarks over general-purpose code-oriented generalization.

In this paper, we propose a more challenging Python code generation benchmark: \textbf{Less} \textbf{Basic} \textbf{Python} \textbf{Problems}. \datasetname is similar in size to \humaneval and \MBPP, but designed to be more complex using model in the loop filtering. \datasetname is also designed to share no inspiration or sources with existing training and evaluation datasets, presenting a novel generalization challenge to contemporary LLMs. In Section \ref{sec:lbpp}, we observe that SOTA models on \humaneval perform up to $43\%$ worse on \datasetname. We contribute \datasetname as a genuinely held-out test to measure \textit{current} code generation capability, and potential overfitting to \humaneval and \MBPP. 

\section{Related Work}
\label{sec:rw}

\humaneval~\citep{humaneval} and \MBPP~\citep{mbpp} remain the most reported results on public leaderboards, but similar datasets exist~\citep{hendrycks2021measuring,li2022competition}.
They consist of short and mostly simple (not programming competition level) instructions with completions in Python.
These datasets have also been translated into more programming languages~\citep{muennighoff2023octopack,cassano2022multipl}, as well as versions with additional tests~\citep{liu2024your}.

\citet{jain2024livecodebench} proposes a continuously updated set of interview questions to improve dataset challenge by including harder and novel (unseen) prompts.
\citet{jimenez2024swebench} aims for challenging software engineering problems requiring understanding of entire repositories.
In a similar vein, RepoQA \citep{repoqa} and Bug In The Code Stack \citep{lee2024bugcodestackllms} focus on understanding long contexts within code tasks. \citet{zhang2024careful} proposes using hidden evaluation sets, however, this approach does not allow inspection of failure cases and requires trusting the quality and correctness of an opaque `black-box' evaluation setup.

Recently, \citet{riddell2024quantifying} analyzed data contamination in popular pretraining datasets:~reporting that  $12.2\%$ of \humaneval samples are present in The Pile~\citep{gao2020pile}, and $18.9\%$ in The Stack~\citep{kocetkov2022stack}. While \citet{riddell2024quantifying} reports ``we do not find the performance rankings of the models to change with decontaminated results'', we identify the ranking between models to vary between contaminated and uncontaminated evaluation datasets (see Table \ref{table:results}). 

\section{Possible sources of contamination}
\label{sec:contamination}
We provide three hypotheses, with respective evidence, on why existing evaluation datasets are leaked and models may already be over-optimized towards existing leaked benchmarks.

\subsection{Direct data leakage}
\label{sec:contamination:dataleakage}
The most obvious reason is the simplest: many of the test datasets are of widespread use and the simplest answer is that modern LLMs are trained on this evaluation data. We note that intentional (i.e., to \textit{cheat}) or unintentional contamination has the same net effect: training on evaluation data limits the confidence and utility of the benchmark.

\begin{figure}[t]
    \centering
    \includegraphics[width=\columnwidth]{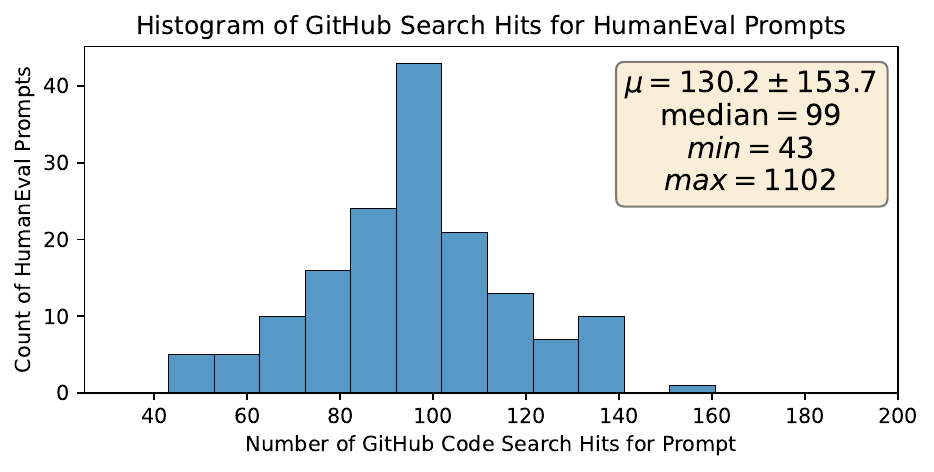}
    \caption{Histogram (excluding outliers) of occurrences for \humaneval~prompts in public GitHub repositories. Every prompt occurs at least $43$ times.}
    \vspace{-2em}
    \label{fig:github_hits}
\end{figure}

Curating high-quality datasets of natural language to code instructions can incur exorbitant costs when one example may cost upwards of dozens of US dollars.
For any party considering the Pareto optimality of dataset size and coding performance gain, the required funding to create novel data can quickly explode.
This leads to a common practice of web scraping code-oriented resources (e.g., GitHub or Stack Overflow) for data. 
However, these resources are also likely sources of contamination. 
The small data size and portability of such benchmarks encourages replication.
We demonstrate this proliferation by keyword searching for \humaneval prompts on GitHub. Fig.~\ref{fig:github_hits} shows that we return a hit in all cases---the median hits is $99$ and the minimum $43$. These hits are often exact duplicates indicating a fork of the original dataset.

While decontamination of training sets is becoming more common, present decontamination filters designed for natural text adapt poorly to code.
To operate efficiently at scale, most filters rely on generic deduplication algorithms e.g., such as $n$-gram matching or hashing functions~\cite{lee-etal-2022-deduplicating}. 
Such surface-level matching does not adequately capture code similarity where a simple variable name change leaves program semantics unchanged, but changing a single keyword can have profound changes.\footnote{Compare the instruction ``\texttt{return true if string is float}'' with ``\texttt{return true if string is verb}''.}
As an example,~\citet{elazar2024whats} report that only $1.22\%$ of verbatim \humaneval is present in the OSCAR popular web corpus.
The same shortcomings of decontamination apply to the creation of large-scale synthetic datasets: for example, the model-generated dataset of Starcoder~\citep{li2023starcoder} is decontaminated only by removing exact docstrings or solutions that match \humaneval or \MBPP.

The recent exploration of~\citet{riddell2024quantifying} quantifies the proportion of this data leakage in existing datasets using plagiarism tools specifically designed for code. 
Even when static training datasets are cleaned, contamination may persist through incremental leakage in other sources.
For example, entities serving models via an API may encounter these benchmark examples when evaluated by third-party users. When a sample of real model usage is annotated for future training data, samples from benchmark evaluation can leak into future training corpora. Furthermore, these samples may include paraphrases and format changes that further complicate heuristic deduplication. 
In this scenario, a model may easily memorize completions to purportedly novel prompts. As evidence of this phenomenon, we prompted one popular commercial system (kept anonymous) with partial prompts from \humaneval that were designed to keep the instruction under-specified.
Table~\ref{table:leakage} in Appendix \ref{appendix1} shows the outcome and evidence that---despite the ambiguity of the prompt---the result matches exactly the gold solution from \humaneval.

\begin{figure}[t]
    \centering
    \includegraphics[width=\columnwidth]{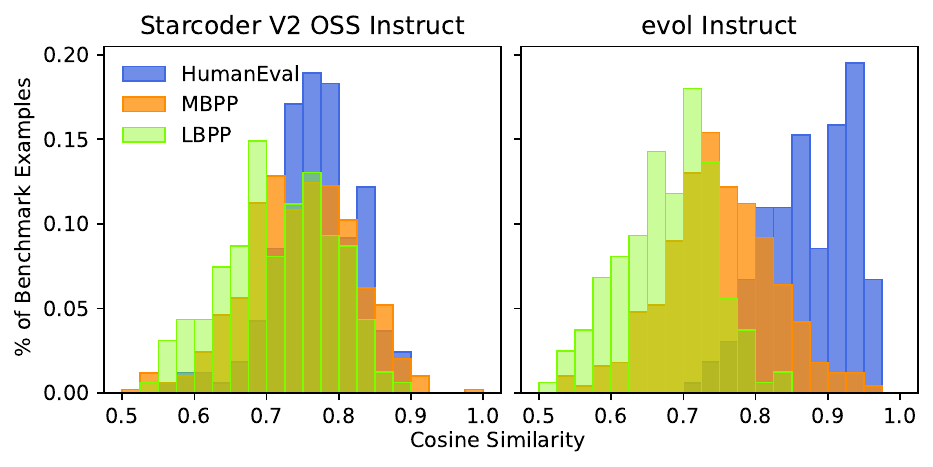}
    \caption{Histogram of cosine similarities for prompts in \humaneval, \MBPP and \datasetname relative to two popular synthetic code training datasets. We note the high similarity between most \humaneval prompts to \textit{evol-instruct}, and how \datasetname has reduced overall similarity to either training dataset. 
    }\label{fig:main}
    \vspace{-1em}
\end{figure}

\subsection{Data leakage through synthetic data}
\label{sec:contamination:synth}

The most capable of code language models rely heavily on the use of synthetic training data~\citep{wizardcoder,magicoder,StarCoder2Instruct,llama3}.
A typical pipeline consists of curating prompts related to code generation, inferring completions with a previously trained LLM, and synthesizing unit tests for relevant prompts using LLMs.
Completions passing the respective unit tests are considered valid code solutions and can be used as future training examples.
Alternatively, if a sufficiently powerful model is used, completions might be used as-is without validation.

Synthetic data unlocks scales that are usually not reachable with human-labeled data. Common synthetic code datasets generally have between tens of thousands and millions of examples.
For example, \textit{Starcoder2-Instruct} is a code dataset of around 238k instances (prior to deduplication) that was created by sampling code from GitHub and using it as seeds to generate self-contained code problems, solutions, and associated tests. \texttt{evol-instruct} is another widely popular dataset used by code LLMs such as WizardCoder~\citep{wizardcoder} for training. It comprises $110k$ complex query prompts with non-verified completions from closed and open-source models.\footnote{Per downloads, the most popular version is a `lightly decontaminated' version \href{https://huggingface.co/datasets/ise-uiuc/Magicoder-Evol-Instruct-110K}{on HuggingFace here}.}
The sheer size of these datasets -- compared to the domain they are targeting -- might explain some memorization.
After all, the number of unique, self-contained interview-like prompts with a reasonable size is fairly limited, and it is possible that synthetic datasets cover a majority of this space. When training data covers most examples of a domain, it does not matter whether the model memorizes the training data or whether it can generalize further.
Table \ref{table:duplicates} shows some examples of very similar (but not necessarily equivalent) data between \texttt{evol-instruct} and \MBPP.

However, the use of a synthetic data pipeline might hide real leakage.
Prior reports ~\citep[page 8]{yu2023wavecoder},~\citep[page 4]{magicoder} discuss an apparent high similarity between examples in \texttt{evol-instruct} and \humaneval. We also found a lot of semantically equivalent prompts between these two datasets and displayed some in Table \ref{table:duplicates_evolinstruct}.
We extend this analysis by studying the similarity between \textit{embedded representations}\footnote{Embedded using Cohere \texttt{embed v3}~\citep{embedv3}.} of the prompts of \humaneval and \MBPP with nearest neighbors from \texttt{evol-instruct} and \textit{Starcoder-Instruct}. Fig.~\ref{fig:main} highlights a widespread similarity between synthetic training datasets and public evaluation data, while the similarity with \datasetname is uniformly lower. This is despite \datasetname having a very similar format to \MBPP and \humaneval (short prompts asking to solve logic problems). The main difference between \MBPP / \humaneval and \datasetname is the difficulty level: \datasetname's questions are generally harder (see Section \ref{sec:lbpp}). This inference aligns with observations when fine-tuning a `Command R Refresh' model adding \texttt{evol-instruct}. \humaneval score increases by +9\%, \MBPP increases by +2\%, but the \datasetname score is unchanged.

Whether the high similarity of synthetic datasets with public evaluation data is due to synthetic data filling the space of problems similar to \humaneval / \MBPP or more direct leakage (eg, through the use of in-context examples), these results point to a larger issue. \humaneval / \MBPP cannot be used as the only proxies to evaluate a model's code abilities. They mostly provide code performance signal on a very specific type of problems with a very specific level of difficulty. We need more diversity in the code evaluation benchmarks and we believe that \datasetname is a step in the right direction.

\subsection{Overfitting to test sets}
\label{sec:contamination:overfitting}
The exaggerated importance of a few benchmarks encourages an incentive structure where model selection prioritizes gain on a narrow suite of metrics. While it may be tempting to use such benchmarks as a deciding factor between similar checkpoints, section \ref{sec:contamination:synth} shows that the correlation between these benchmarks and `solving code generation' is weak. While the meaning and measurement of this unscientific objective are subject to constant revision, selecting for optimal \humaneval performance is akin to $p$-hacking in other fields.
This can be justified by assuming these benchmarks are the new dev sets, while the true test is the usage of users over time.
However, the usefulness of a dev set entirely relies on its similarity with the actual use case. 

Moreover, some risk remains that models overfit to these `lucrative' benchmarks, distorting the perception of downstream performance. Table \ref{table:results} and Fig.~\ref{fig:correlation} illustrate this problem well. Even though the correlation between \MBPP/\humaneval scores and \datasetname scores is strong, some models are ranking noticeably higher on \MBPP/\humaneval than on \datasetname.

The ultimate effect is imbalanced optimization solely towards these metrics, further motivating the practices outlined in Sections \ref{sec:contamination:dataleakage} and \ref{sec:contamination:synth}.

\input{tables/lbpp_examples}

\section{\datasetname:~Less Basic Python Problems}
\label{sec:lbpp}

As mentioned above, we have created \textbf{Less Basic Python Problems (\datasetname)} for a less biased measure of modern models' code capabilities. \datasetname is a dataset of 161 code completion problems in the style of \humaneval. All prompts include evaluation unit tests, with a median of 4 tests per example.

\paragraph{Dataset Annotation:}
Human annotators were asked to create brand new problems that were not solvable by a strong internal model in the loop.
All annotators had competitive programming experience. They were instructed to come up with unique problems either from scratch or inspired by programming textbooks whose content was not freely available (e.g., searchable or unlicensed) on the Internet. Annotators could not copy any exercise from a web source or any LLM and only use these sources for inspiration. All sources were cited by annotators and prompts were verified to not match source inspiration. Every prompt is also manually verified as not easily searchable on the web at the time of writing. Each prompt went through additional review to ensure that they were original, hard, and unambiguous. Around a third of the suggested prompts were disqualified for one of these reasons. 

All annotators were paid above minimum wage in their respective countries, and all final prompt-completion pairs were manually reviewed by the authors. 
This adversarial collection resulted in more difficult problems, with most models solving less than $50\%$ of the dataset.

\input{tables/results}

\begin{figure}[t]
    \centering
    
    \begin{subfigure}{0.5\textwidth}
        \caption{\humaneval}
        \includegraphics[width=\linewidth]{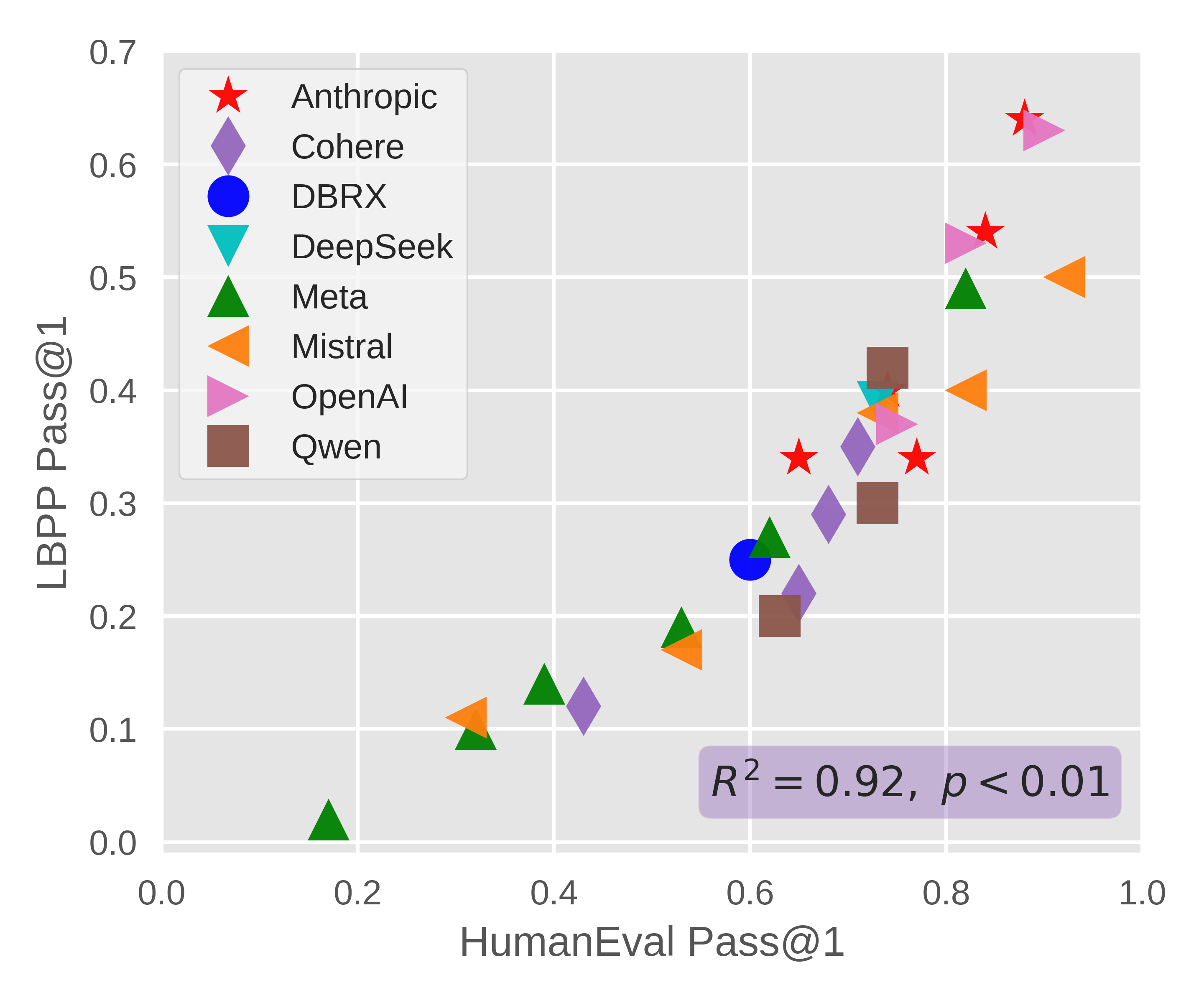} 
        \vspace{-2em}
        \label{fig:correlation_he}
    \end{subfigure}
    
    \begin{subfigure}{0.5\textwidth}
        \caption{\MBPP}
        \includegraphics[width=\linewidth]{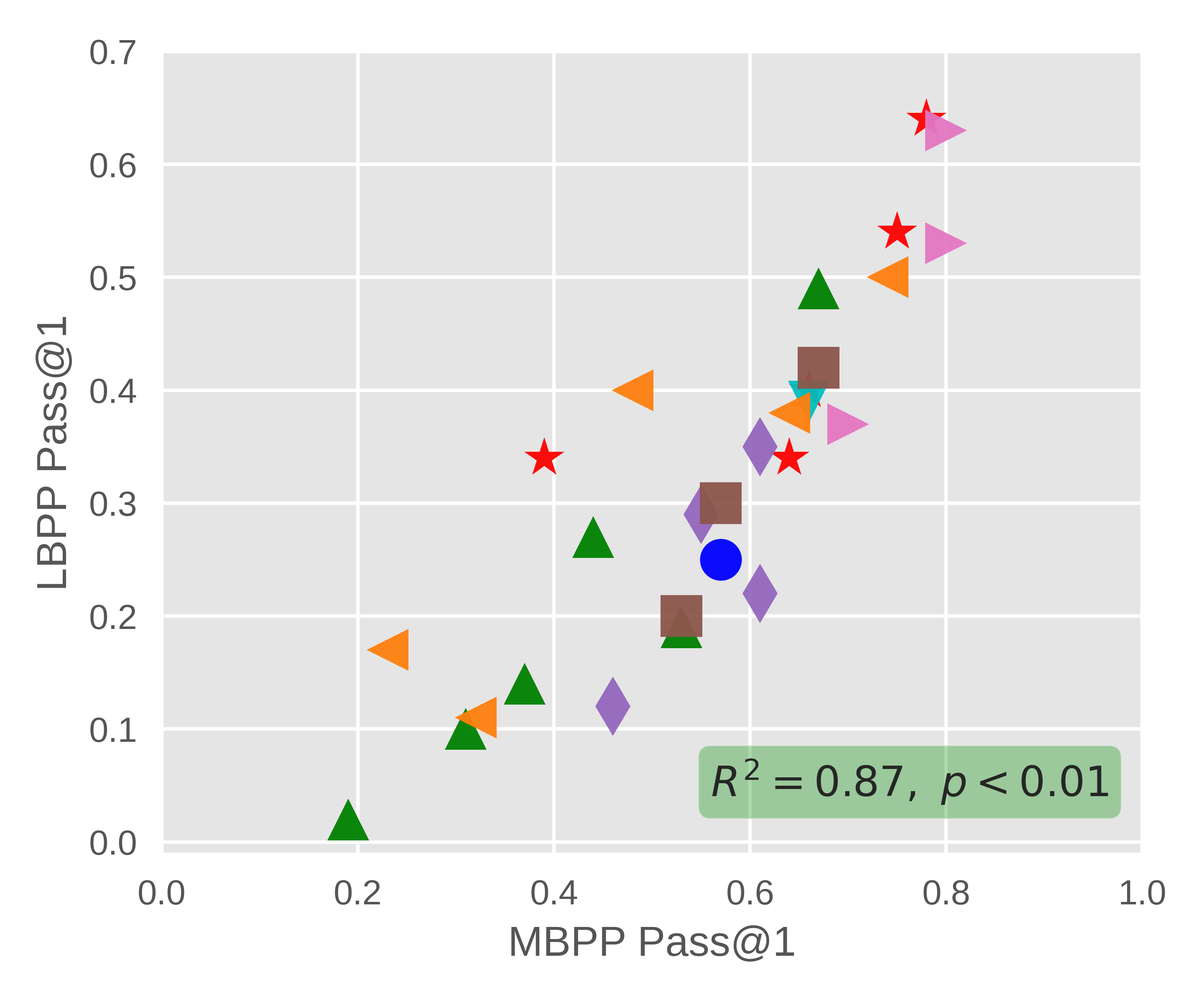}         
        \vspace{-2em}
        \label{fig:correlation_mbpp}
    \end{subfigure}
    
    \caption{\small Pass@1 rate of \datasetname against (a) \humaneval and  (b) \MBPP. \datasetname performance correlates with both prior datasets, but is designed to be genuinely unseen by contemporary LLMs.}
    \vspace{-2em}
    \label{fig:correlation}
\end{figure}

\paragraph{Initial Results:}
Table~\ref{table:results} shows Pass@1 on LBPP for a range of models. Notably, leading models for \humaneval and \MBPP perform up to $43\%$ and $27\%$ poorer on \datasetname respectively. Additionally, model rankings between either \humaneval and \MBPP update for \datasetname, potentially identifying overfitting to public benchmarks when presented with a challenging, unseen test set. 
\datasetname is a similarly reliable indicator of code generation performance than prior benchmarks. In Fig.~\ref{fig:correlation}, we observe a strong significant correlation between `Pass@1' scores of either \humaneval or \MBPP and \datasetname. 
While existing public benchmarks are still a valuable target signal for performance, \datasetname is additionally advantageous in that the problems are harder (see Table~\ref{table:results}), the dataset is uncontaminated in current training corpora, and prompts bear lower resemblance to existing synthetic corpora (see Fig.~\ref{fig:main}). 

\paragraph{Challenges in \datasetname:} We study common errors and mistakes from multiple models to identify the most challenging features in \datasetname. Examples of problems unsolved by all models are shown in Table \ref{tab:lbpp_examples}. Considering common errors between Claude 3.5-Sonnet and Command R Refresh, as the best and recently released model respectively, we identify multiple core trends in failure. Of mutual errors: 21\% are related to problems on 2D and 3D arrays; 18\% are related to graph-oriented algorithms; and 17\% are concerning complex programming concepts often used in competition settings. Additional challenging topics include bit arithmetic \& manipulation (8\%), Pandas data processing  (8\%), and file IO (8\%). The range of shortcomings between all models highlights the variety of domains that future LLMs must master to improve code generation. Overall, the diversity and difficulty of the problems in \datasetname challenges even purportedly advanced models with novel and unsolved prompts.

\section{Conclusion}

We study the cause and effect of data contamination via two popular code generation benchmarks. Our analysis highlights that contamination is likely unavoidable at the LLM scale given the difficulty of filtering every potential permutation of a benchmark dataset. 
This insight motivates our contribution of \datasetname:~a novel code generation benchmark to evaluate contemporary LLMs in a contamination-free setting. 
We are well aware that our decision to release this dataset will make future leakage impossible to control.
However, with the context of the fast-paced model development cycles that LLMs are currently undergoing we believe that releasing this increases the trustworthiness and usefulness of this dataset.
It is conveniently designed to serve as a drop-in replacement (or addition) of current evaluation sets.
On top of its novelty, the more challenging nature of this dataset also provides a cleaner signal for model comparison.

\section{Limitations}
All the model analysis was done \textit{black-box}, without inspecting the model weights or the training set (except the work on synthetic data).
There is no reason why this dataset will not follow the same path as the two studied here.
As mentioned in the Conclusion we believe there is more value in that than in an alternative solution (not releasing or keeping it behind an API access).

\bibliography{anthology,custom}
\bibliographystyle{acl_natbib}

\appendix
\section{Appendix}
\label{appendix1}

\input{tables/duplicates}

\begin{landscape}
    \begin{table}
        \centering
        \caption{Original human evaluation prompts with the completion from a major LLM provider.}
        \label{table:leakage}
        \input{tables/leakage}
    
    \end{table}
\end{landscape}

\begin{landscape}

    \begin{table}
            \centering
            \caption{Most similar prompt in \texttt{evol-instruct} for a random sample of \humaneval prompts.}
            \label{table:evol_he}
            \input{tables/leakage_evol_he}
            
         \label{table:duplicates_evolinstruct}
    \end{table}
\end{landscape}

\end{document}

%% file: tables/lbpp_examples.tex
\begin{table}[t]
\centering
{
\footnotesize
\begin{tabular}{@{}l@{}}
\toprule
Unsolved problems in \datasetname \\ \midrule
\begin{tabular}[c]{@{}l@{}}Given a list of unique words each of size $k$ \\ and an $n$ sized word, w, where n is a multiple of $k$,\\ write a program in Python to determine the number \\ of unique combinations of words in the list that can \\ be concatenated to form an anagram of the word.\end{tabular} \\ \midrule
\begin{tabular}[c]{@{}l@{}}Write a function in python that takes as input a \\ recursive function and some parameters. The function \\ should return the number of times the recursive function \\ ran itself when starting it with the provided parameters.\end{tabular} \\ \midrule
\begin{tabular}[c]{@{}l@{}}Write a Monte Carlo function in Python to compute \\ the median number of cards you'd need to draw \\ from a deck such that the sum equals or exceeds \\ the value $V$.\end{tabular} \\ \bottomrule
\end{tabular}%
}
\caption{Sampled prompts in \datasetname unsolved by leaders on existing benchmarks. Prompts shortened for brevity.}
\label{tab:lbpp_examples}
\vspace{-1.5em}
\end{table}

%% file: tables/results.tex
\begin{table}[t!]
\resizebox{\columnwidth}{!}{%
\begin{tabular}{@{}clcccc@{}}
\toprule
& Model Name                  & \humaneval & \MBPP & \datasetname & HumanEval$\rightarrow$\datasetname \\ \midrule
\multirow{5}{*}{\STAB{\rotatebox[origin=c]{90}{Mistral}}}
& Mistral 7B                  & 0.31 & 0.32 & 0.11 & 27$\rightarrow$26 \\ 
& Mixtral $8\times7$B         & 0.53 & 0.23 & 0.17 & 22$\rightarrow$23 \\
& Mixtral $8\times22$B        & 0.73 & 0.64 & 0.38 & 13$\rightarrow$11 \\
& Mistral Large               & {\bf 0.92} & 0.74 & 0.50 & 1$\rightarrow$5   \\ 
& Codestral 22B               & 0.82 & 0.48 & 0.40 & 5$\rightarrow$9  \\ 
\midrule
\multirow{6}{*}{\STAB{\rotatebox[origin=c]{90}{Meta}}}
& Codellama 7B Instruct       & 0.39 & 0.37 & 0.14 & 25$\rightarrow$24 \\
& Codellama 34B Instruct      & 0.53 & 0.53 & 0.19 & 23$\rightarrow$22 \\
& Llama2 7B Chat              & 0.17 & 0.19 & 0.02 & 28$\rightarrow$28 \\
& Llama2 60B Chat             & 0.32 & 0.31 & 0.10 & 26$\rightarrow$27 \\
& Llama3 8B Instruct          & 0.62 & 0.44 & 0.27 & 20$\rightarrow$18 \\
& Llama3 70B Instruct         & 0.82 & 0.67 & 0.49 & 6$\rightarrow$6   \\ 
\midrule
\multirow{3}{*}{\STAB{\rotatebox[origin=c]{90}{OpenAI}}}
& GPT-3.5 Turbo               & 0.75 & 0.70 & 0.37 & 9$\rightarrow$12  \\
& GPT-4                       & 0.82 & {\bf 0.80} & 0.53 & 7$\rightarrow$4   \\
& GPT-4o                      & 0.90 & {\bf 0.80} & 0.63 & 2$\rightarrow$2   \\ 
\midrule
\multirow{5}{*}{\STAB{\rotatebox[origin=c]{90}{Antropic}}}
& Claude-2                    & 0.65 & 0.39 & 0.34 & 17$\rightarrow$15 \\
& Claude-3-Haiku              & 0.77 & 0.64 & 0.34 & 8$\rightarrow$14 \\
& Claude-3-Sonnet             & 0.74 & 0.66 & 0.40 & 10$\rightarrow$8  \\
& Claude-3-Opus               & 0.84 & 0.75 & 0.54 & 4$\rightarrow$3   \\
& Claude-3.5-Sonnet           & 0.88 & 0.78 & {\bf 0.64} & 3$\rightarrow$1   \\ 
\midrule
\multirow{3}{*}{\STAB{\rotatebox[origin=c]{90}{Qwen}}}
& Qwen1.5 72B Chat            & 0.63 & 0.53 & 0.20 & 19$\rightarrow$21 \\
& Qwen1.5 110B Chat           & 0.73 & 0.57 & 0.30 & 12$\rightarrow$16 \\
& Qwen 2 72B Instruct         & 0.74 & 0.67 & 0.42 & 11$\rightarrow$7   \\ 
\midrule
\multirow{4}{*}{\STAB{\rotatebox[origin=c]{90}{Cohere}}}
& Command R                   & 0.43 & 0.46 & 0.12 & 24$\rightarrow$25 \\
& Command R (Refresh)         & 0.71 & 0.55 & 0.35 & 15$\rightarrow$13 \\ 
& Command R+                  & 0.65 & 0.61 & 0.22 & 18$\rightarrow$20 \\
& Command R+ (Refresh)        & 0.68 & 0.61 & 0.29 & 16$\rightarrow$17 \\ 
\midrule
& Deepseek Coder 33B Instr.   & 0.73 & 0.66 & 0.39 & 14$\rightarrow$10  \\  
\midrule
& Databricks DBRX Instr.      & 0.60 & 0.57 & 0.25 & 21$\rightarrow$19 \\
\bottomrule
\end{tabular}%
}
\caption{Pass@1 results across popular models for zero-shot \humaneval, \MBPP and \datasetname. All models perform worse on \datasetname than either existing benchmark. Rankings between models \textbf{change} between \humaneval and \datasetname, contrasting to \citet{riddell-etal-2021-call}. Model rankings
similarly also change between \MBPP and \datasetname.\label{table:results}}
\vspace{-1em}
\end{table}

%% file: tables/duplicates.tex

\begin{table}[h]
\resizebox{\columnwidth}{!}{%
\begin{tabular}{@{}ll@{}}
\toprule
\textbf{MBPP test prompt}                                                                                      & \textbf{\texttt{evol-instruct} prompt} \\ \midrule
\begin{tabular}[c]{@{}l@{}}Write a function to find the\\  perimeter of a square\end{tabular}                  & Calculate the perimeter of a square    \\ \midrule
\begin{tabular}[c]{@{}l@{}}Write a python function to find\\  the volume of a triangular prism\end{tabular} &
  \begin{tabular}[c]{@{}l@{}}How can I use JavaScript to \\ calculate the volume of a \\ triangular prism with an irregular \\ base and uneven side lengths?\end{tabular} \\ \midrule
\begin{tabular}[c]{@{}l@{}}Write a function to convert snake\\  case string to camel case string.\end{tabular} & Convert the sentence to camel case.    \\ \bottomrule
\end{tabular}%
}
\caption{Similarity in prompts between \MBPP evaluation dataset and \texttt{evol-instruct} synthetic training dataset.}
\label{table:duplicates}
\end{table}

%% file: tables/leakage.tex
{\scriptsize

\begin{tabular}{p{8cm}|p{8cm}|p{8cm}}
\toprule
\textbf{Original human-eval prompt} & \textbf{Submitted Prompt} & \textbf{Opus completion} \\ \midrule
\begin{lstlisting}[basicstyle=\ttfamily\tiny]
def is_happy(s):
"""You are given a string s.
Your task is to check if the string is happy or not.
A string is happy if its length is at least 3
and every 3 consecutive letters are distinct
For example:
is_happy(a) => False
is_happy(aa) => False
is_happy(abcd) => True
is_happy(aabb) => False
is_happy(adb) => True
is_happy(xyy) => False """
\end{lstlisting} & 
Continue this code. Do not add any test, just continue it: \newline
\begin{lstlisting}[basicstyle=\ttfamily\tiny]
def is_happy(s):
""" You are given a string s. Your task is to check if the str
\end{lstlisting}
& 
Here’s the continuation of the code without adding any tests:
\begin{lstlisting}[basicstyle=\ttfamily\tiny]
def is_happy(s):
"""
You are given a string s. Your task is to check if the string 
is happy. A string is called happy if its length is at least 3 
and every 3 consecutive letters are distinct

[...]
\end{lstlisting}
\\ \midrule

\begin{lstlisting}[basicstyle=\ttfamily\tiny]
def count_upper(s):
""" Given a string s, count the number of uppercase vowels
in even indices.

For example:
count_upper('aBCdEf') returns 1
count_upper('abcdefg') returns 0
count_upper('dBBE') returns 0
"""
\end{lstlisting} & Complete this instruction, that is taken from the Human Eval test set: 
\begin{lstlisting}[basicstyle=\ttfamily\tiny]

def count_upper(s):
""" Given a string s, count the number of
\end{lstlisting} & 
Here is the completion of the given function signature and docstring to count the number of uppercase letters in a string:
\begin{lstlisting}[basicstyle=\ttfamily\tiny]
def count_upper(s):
"""
Given a string s, count the number of uppercase vowels in 
the string.

[...]
\end{lstlisting}
\\ \midrule

\begin{lstlisting}[basicstyle=\ttfamily\tiny]
def any_int(x, y, z):
''' Create a function that takes 3 numbers.
Returns true if one of the numbers is equal to the sum of the 
other two, and all numbers are integers. Returns false in any 
other cases.

Examples:
any_int(5, 2, 7) -> True
any_int(3, 2, 2) -> False
any_int(3, -2, 1) -> True
any_int(3.6, -2.2, 2) -> False '''
\end{lstlisting} & 
continue this code.
\begin{lstlisting}[basicstyle=\ttfamily\tiny]
def any_int(x, y, z): 
"""Create a function that takes 3 numbers. Return true i
\end{lstlisting} & 
\begin{lstlisting}[basicstyle=\ttfamily\tiny]
def any_int(x, y, z):
"""
Create a function that takes 3 numbers.
Returns true if one of the numbers is equal to the sum of the 
other two numbers. Otherwise returns false.
"""
[...]
\end{lstlisting} \\

\midrule
\begin{lstlisting}[basicstyle=\ttfamily\tiny]
def order_by_points(nums):
""" Write a function which sorts the given list of integers
in ascending order according to the sum of their digits.

Note: if there are several items with similar sum of their 
digits, order them based on their index in original list.

For example:

order_by_points([1, 11, -1, -11, -12]) == [-1, -11, 1, -12, 11]
order_by_points([]) == []
"""
\end{lstlisting} & 
You will be given an instruction to complete a python function. This instruction is cut in half, so you need to complete it. No need of generating the final code, only the instruction is required.

The incomplete instruction is:

\begin{lstlisting}[basicstyle=\ttfamily\tiny]
def order_by_points(nums): 
""" Write a function which sorts the given list of integers in 
ascending order according
\end{lstlisting} & 
Here's the completed instruction for the Python function:

\begin{lstlisting}[basicstyle=\ttfamily\tiny]
def order_by_points(nums):
""" Write a function which sorts the given list of integers in 
ascending order according to the sum of their digits.
If two integers have the same sum of their digits, order them 
based on their index in the original list.
"""
\end{lstlisting} \\
\bottomrule
\end{tabular}

}

%% file: tables/leakage_evol_he.tex
{\scriptsize

\begin{tabular}{p{12cm}|p{12cm}}
\toprule
\textbf{\humaneval prompt} & \textbf{Closest \texttt{evol-instrct} prompt} \\ \midrule
\begin{lstlisting}[basicstyle=\ttfamily\tiny]
Write a Python function `longest(strings: List[str]) -> Optional[str]` to solve the following 
problem:
Out of list of strings, return the longest one. Return the first one in case of multiple
strings of the same length. Return None in case the input list is empty.
>>> longest([])
>>> longest(['a', 'b', 'c'])
'a'
>>> longest(['a', 'bb', 'ccc'])
'ccc'
\end{lstlisting} & 
\begin{lstlisting}[basicstyle=\ttfamily\tiny]
Complete the code below, considering an augmented depth of inquiry and maintaining brevity:
from typing import List, Optional

def longest(strings: List[str]) -> Optional[str]:
    """ From a list of strings, return the longest one. For multiple strings with equal
    length, return the first. For an empty list, return None.
    >>> longest([])

    >>> longest(['a', 'b', 'c'])
    'a'
    >>> longest(['a', 'bb', 'ccc'])
    'ccc'
\end{lstlisting}
\\ \midrule

\begin{lstlisting}[basicstyle=\ttfamily\tiny]
Write a Python function `make_a_pile(n)` to solve the following problem:
Given a positive integer n, you have to make a pile of n levels of stones.
The first level has n stones.
The number of stones in the next level is:
- the next odd number if n is odd.
- the next even number if n is even.
Return the number of stones in each level in a list, where element at index
i represents the number of stones in the level (i+1).
Examples:
>>> make_a_pile(3)
[3, 5, 7]
\end{lstlisting} &

\begin{lstlisting}[basicstyle=\ttfamily\tiny]
Please complete the following code with added difficulty:

def make_a_pile(n, pattern):
    """
    Given a positive integer n, you have to make a pile of n levels of stones.
    The first level has n stones. The number of stones in the next level is determined
    by the given pattern 'odd' or 'even':
        - If pattern is 'odd', add the next odd number to the previous level stones.
        - If pattern is 'even', add the next even number to the previous level stones.
    Return the number of stones in each level in a list, where element at index
    i represents the number of stones in the level (i+1).

    Examples:
    >>> make_a_pile(3, 'odd')
    [3, 5, 7]
    >>> make_a_pile(3, 'even')
    [3, 6, 9]
    """
\end{lstlisting} 
\\ \midrule

\begin{lstlisting}[basicstyle=\ttfamily\tiny]
Write a Python function `x_or_y(n, x, y)` to solve the following problem:
A simple program which should return the value of x if n is
a prime number and should return the value of y otherwise.
Examples:
for x_or_y(7, 34, 12) == 34
for x_or_y(15, 8, 5) == 5
\end{lstlisting} & 
\begin{lstlisting}[basicstyle=\ttfamily\tiny]
Complete the subsequent lines of programming code:
/*
An elementary program intended to, given a single n value, decipher between two 
distinct possibilities:
- If the provided n is a numerical prime, the program should yield the 
output value equivalent to variable x.
- Should the n fail to meet the prime criteria, the program should then 
present the value contained within variable y as its output.

Examples:
Upon running x_or_y(7, 34, 12) the output will be 34
While executing x_or_y(15, 8, 5) the output will be 5

*/
#include<stdio.h>
using namespace std;
int x_or_y(int n,int x,int y){
\end{lstlisting} \\
\bottomrule
\end{tabular}

}